\documentclass[letterpaper, 10 pt, conference]{ieeeconf}  % Comment this line out if you need a4paper

\IEEEoverridecommandlockouts                              % This command is only needed if 
\usepackage{graphicx} % for pdf, bitmapped graphics files
\usepackage{amsmath} % assumes amsmath package installed
\usepackage{amssymb}  % assumes amsmath package installed
\usepackage{newtxmath}
\usepackage[space]{cite}
\usepackage{bm}
\usepackage{booktabs}
\usepackage{tabularx}
\usepackage{color}
\usepackage{url}
\usepackage{multirow}
\usepackage{gensymb}
\usepackage{dsfont}

\newcolumntype{C}{>{\centering\arraybackslash}X}
\newcolumntype{L}{>{\raggedright\arraybackslash}X}
\newcolumntype{R}{>{\raggedleft\arraybackslash}X}
\newcolumntype{P}[1]{>{\centering\arraybackslash}p{#1}}

\title{\LARGE \bf
Learning to Drop Points for LiDAR Scan Synthesis
}

\author{Kazuto Nakashima and Ryo Kurazume% <-this % stops a space
\thanks{Kazuto Nakashima and Ryo Kurazume are with Faculty of Information Science and Electrical Engineering, Kyushu University, 744 Motoka Fukuoka, Japan. {\tt\small k\_nakashima@irvs.ait.kyushu-u.ac.jp}, {\tt\small kurazume@ait.kyushu-u.ac.jp}}%
}

\begin{document}

\maketitle
\thispagestyle{empty}
\pagestyle{empty}

%%%%%%%%%%%%%%%%%%%%%%%%%%%%%%%%%%%%%%%%%%%%%%%%%%%%%%%%%%%%%%%%%%%%%%%%%%%%%%%%
\begin{abstract}
	3D laser scanning by LiDAR sensors plays an important role for mobile robots to understand their surroundings. Nevertheless, not all systems have high resolution and accuracy due to hardware limitations, weather conditions, and so on. Generative modeling of LiDAR data as scene priors is one of the promising solutions to compensate for unreliable or incomplete observations. In this paper, we propose a novel generative model for learning LiDAR data based on generative adversarial networks. As in the related studies, we process LiDAR data as a compact yet lossless representation, a cylindrical depth map. However, despite the smoothness of real-world objects, many points on the depth map are dropped out through the laser measurement, which causes learning difficulty on generative models. To circumvent this issue, we introduce measurement uncertainty into the generation process, which allows the model to learn a disentangled representation of the underlying shape and the dropout noises from a collection of real LiDAR data. To simulate the lossy measurement, we adopt a differentiable sampling framework to drop points based on the learned uncertainty. We demonstrate the effectiveness of our method on synthesis and reconstruction tasks using two datasets. We further showcase potential applications by restoring LiDAR data with various types of corruption.
\end{abstract}

%%%%%%%%%%%%%%%%%%%%%%%%%%%%%%%%%%%%%%%%%%%%%%%%%%%%%%%%%%%%%%%%%%%%%%%%%%%%%%%%
\section{Introduction}

3D scene understanding is indispensable for mobile robots to detect obstacle objects, find traversable paths, and explore the real world. A typical representation of 3D scenes is a point cloud, which can be measured by various range-scanning devices such as 3D LiDARs and RGB-D cameras. In particular, 3D LiDARs are widely used for autonomous driving systems to enable their perception capabilities such as SLAM, object detection, and semantic segmentation.

A 3D LiDAR calculates distances based on the time-of-flight of pulsed laser emitted and reflected at multiple elevation/azimuth angles. However, to gain the angular resolution and precision, it requires many scanner units and sufficient housing size, which could conflict with system requirements. Moreover, the measured points could have noises under adverse weather. The purpose of this study is to reconstruct high-quality point clouds from such limited or corrupted observations. However, it is non-trivial to manipulate a number of raw points with semantic consistency. One promising approach is to learn a generative model of point clouds as scene priors and to find the high-quality sample nearest to the observation.

\begin{figure}[tb]
	\footnotesize
	\centering
	\begin{minipage}[c]{\hsize}
		\centering
		\includegraphics[width=0.9\hsize]{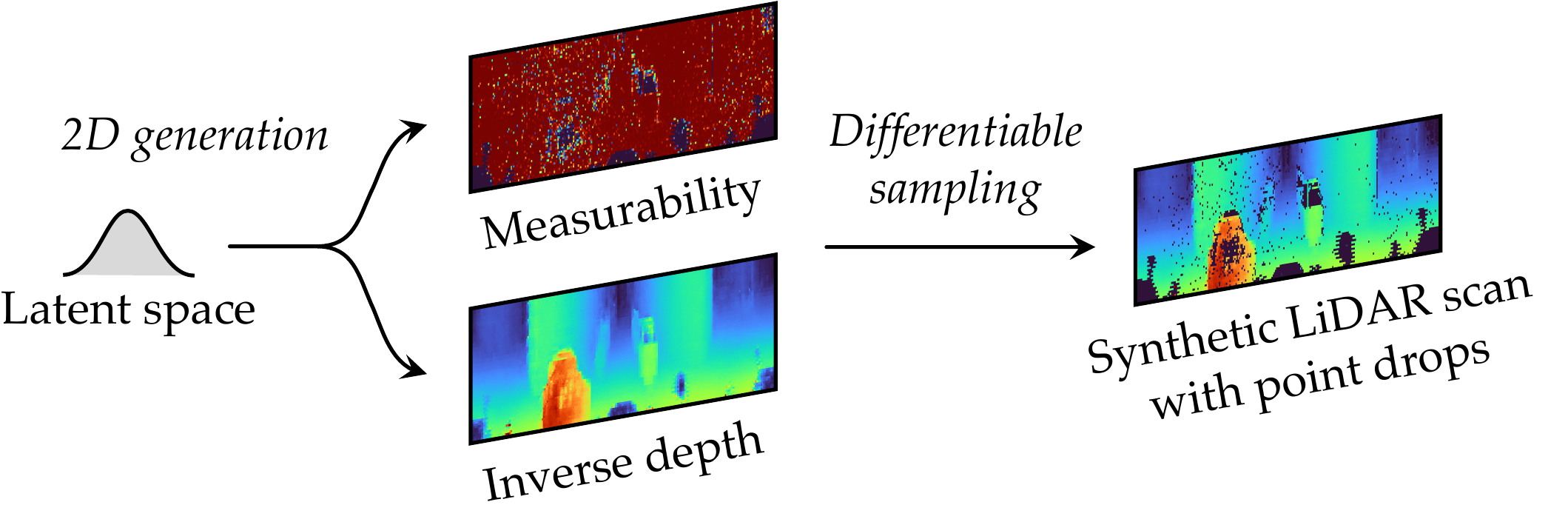}
		\\
		(a) Decomposed synthesis of LiDAR scans
	\end{minipage}
	\\
	\begin{minipage}[c]{\hsize}
		\centering
		\includegraphics[width=\hsize]{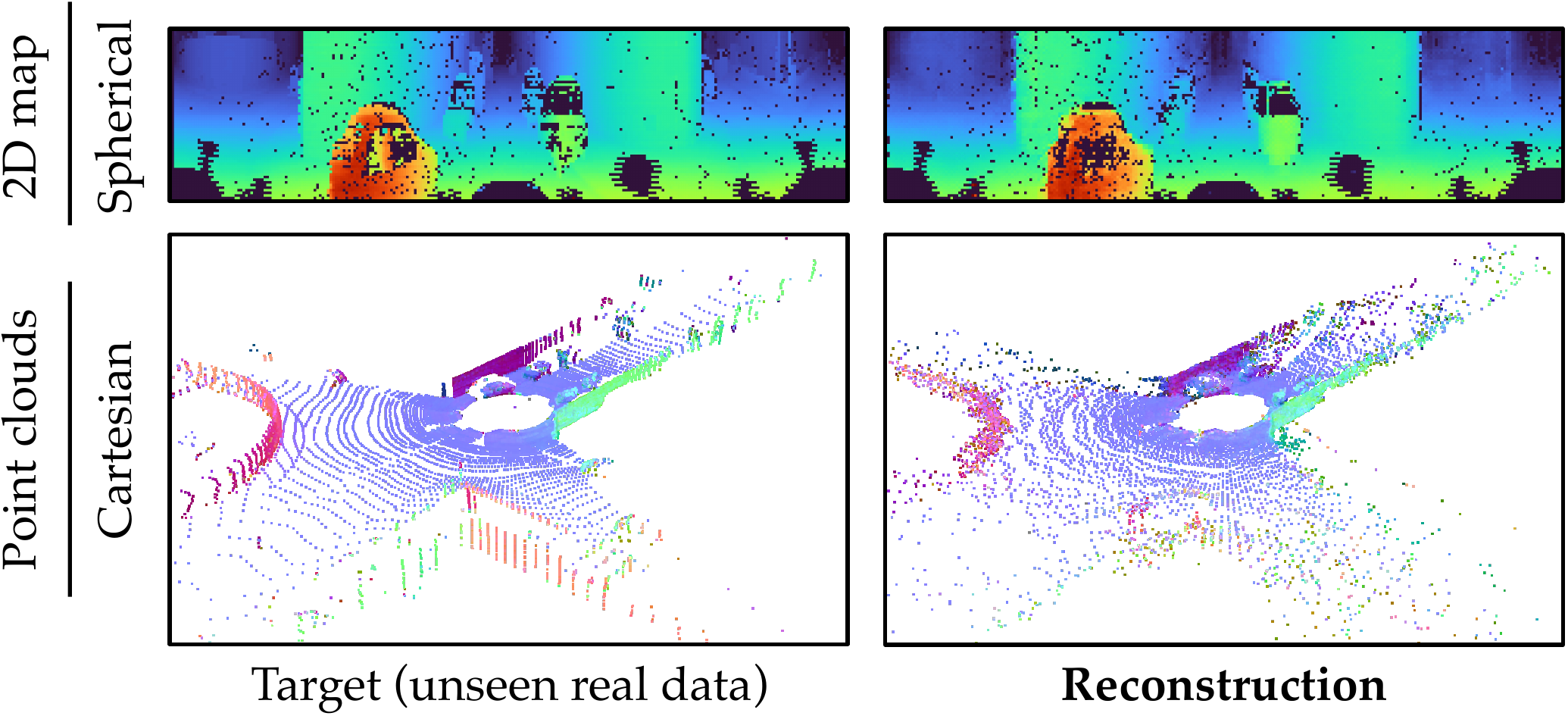}
		\\
		(b) Reconstruction of unseen data from KITTI~\cite{geiger2013vision} test set
	\end{minipage}
	\caption{A concept of our proposed generative model for LiDAR data. (a) With only noisy training data, our proposed method can learn an underlying complete depth map and a measurability map that simulates dropout noises. (b) The trained model can be used to reconstruct unseen data.}
	\label{fig:introduction:summary}
\end{figure}

Generative models have received considerable attention owing to recent advances in representation learning with deep neural networks.
In particular, generative adversarial networks (GANs)~\cite{goodfellow2014generative} are a popular framework for building a generative model.
GANs have been actively studied on 2D tasks such as generating photo-realistic natural images~\cite{karras2020analyzing, karras2018progressive}, while they can also be used to model 3D data such as point clouds.
Several studies~\cite{yang2019pointflow,pmlr-v80-achlioptas18a} proposed generative models that produce point clouds as a set of unordered points and demonstrated on data uniformly sampled from CAD models~\cite{shapenet2015}
In contrast, the point clouds by 3D LiDARs are often large, non-uniform, and sparse, as they are measured by radiated lasers centering on the sensor head.
In other perception tasks with LiDAR data, projection-based 2D representations have also been studied, such as sensor-view depth images by spherical projection~\cite{li2016vehicle,wu2018squeezeseg} and bird's eye view~\cite{zhang2020polarnet}.
The 2D approaches have advantages in computational efficiency: the processing can be designed with 2D convolutional networks, less problem on the sparsity, and the representation is compact yet lossless for spherical projection.
In semantic point segmentation~\cite{wu2018squeezeseg,zhang2020polarnet} and object detection tasks~\cite{li2016vehicle}, these approaches have gained performance in practice.

The goal of this study is to build a generative model of LiDAR data based on the 2D representation~\cite{caccia2019deep,wu2018squeezeseg,zhang2020polarnet,li2016vehicle,manivasagam2020lidarsim}.
In particular, we use the sensor-view representation.
Caccia \textit{et al.}~\cite{caccia2019deep} is closely related to our task, which trained GANs and variational autoencoders (VAEs) with Cartesian points projected onto sensor-view grids.
However, we found that training generative models on this 2D representation is still challenging because of LiDAR-specific noises.
As seen in Fig.~\ref{fig:introduction:summary}(b), the projected image involves scattered dropout noises. The noises are due to the reflection failure of laser measurement, hereinafter called \textit{point-drops}.
Caccia \textit{et al.}~\cite{caccia2019deep} handled this issue by interpolating all the point-drops by neighbor pixels, however, the heuristic imputation could cause unnatural structures.

In this paper, we introduce a novel GAN framework to synthesize \textit{depth} with \textit{uncertainty} from \textit{styles}, named DUSty.
Our key idea is to introduce measurement uncertainty into the generation process so that the underlying complete signals can be modeled implicitly.
As shown in Fig.~\ref{fig:introduction:summary}(a), our model produces a complete depth map and the corresponding measurability to sample realistic \textit{dusty} point-drops.
Our proposed model can be trained only with raw LiDAR data involving point-drops.
We evaluated our method on two LiDAR datasets for driving scenes: KITTI~\cite{geiger2013vision} and MPO~\cite{martinez2019fukuoka}.
We demonstrate that our method successively synthesizes realistic LiDAR data, learning the underlying complete shape.
We also demonstrate the reconstruction capability for unseen real data in various corruption settings.
The code will be available at \url{https://github.com/kazuto1011/dusty-gan}.

The main contributions can be summarized as follows:

\begin{itemize}
	\item We propose a noise-aware GAN framework DUSty, which models a complete depth map and the corresponding measurability from real LiDAR data.
	\item We introduce a differentiable relaxation to learn the discrete distribution of point-drops.
	\item We demonstrate the effectiveness of our approach on synthesis and reconstruction tasks on two LiDAR datasets.
\end{itemize}

\begin{figure*}[tb]
	\centering
	\includegraphics[width=\hsize]{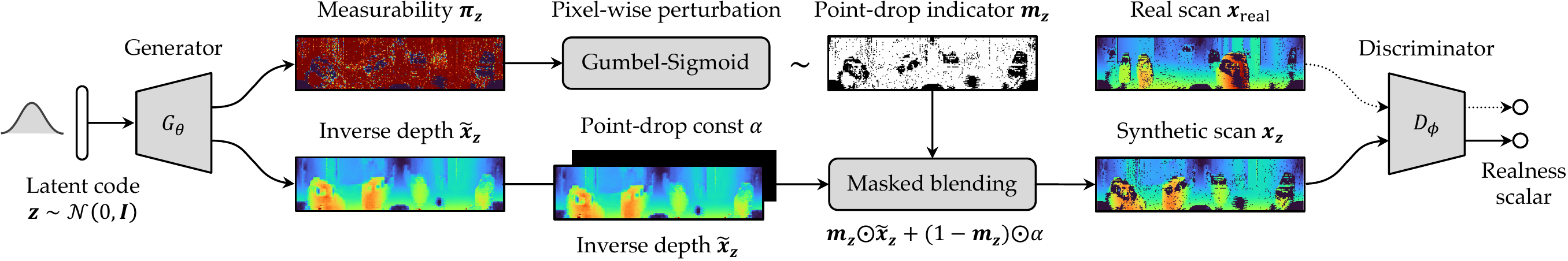}
	\caption{Overview of our proposed GAN framework. Generator $G_{\theta}$ produces the inverse depth map $\tilde{\bm{x}}_{\bm{z}}$ and the corresponding measurability map $\bm{\pi}_{\bm{z}}$, from the sampled latent code $\bm{z}\sim\mathcal{N}(0,\bm{I})$. The binary mask $\bm{m}_{\bm{z}}$ indicating point-drops is sampled from the measurability map $\bm{\pi}_{\bm{z}}$. Discriminator $D_{\phi}$ distinguishes the processed inverse depth map $\bm{x}_{\bm{z}}$ from the real data $\bm{x}_\mathrm{real}$.}
	\label{fig:model}
\end{figure*}

\section{Related Work}

\subsection{Synthesizing LiDAR Data}

Generative modeling of point clouds is a challenging task emerged in recent years.
Most studies focused on small point sets sampled from CAD objects~\cite{yang2019pointflow}, but rarely on LiDAR point clouds.
There have been studies based on simulated environments~\cite{wu2018squeezeseg,manivasagam2020lidarsim}, but their diversity was limited to defined cases.
Only Caccia \textit{et al.}~\cite{caccia2019deep} worked on modeling LiDAR data with two popular frameworks for deep generative models: variational autoencoders (VAEs) and GANs.
The authors revealed that 2D representation was much better for LiDAR data than for point sets.
However, the study evaluated quantitative performance on reconstruction tasks only for VAEs.
In this paper, we provide the results of both quality and diversity metrics for GANs on the synthesis task and demonstrate the improvement in the reconstruction task.
In particular, we found that simulating \textit{point-drops} was important for training GANs effectively.

For obstacle detection tasks, recent studies~\cite{wu2018squeezeseg,manivasagam2020lidarsim} revealed that simulating point-drops could mitigate the domain gap between the real and simulated environments.
Wu \textit{et al.}~\cite{wu2018squeezeseg} calculated the spatial prior from real LiDAR scans to sample dropout noises.
However, the sampled noise is independent of the instance.
Manivasagam \textit{et al.}~\cite{manivasagam2020lidarsim} trained U-Net to predict point-drops as a binary classification task.
However, the drop prediction requires clean LiDAR data superimposed by multiple scans, and the classification probability is not calibrated.
In contrast, our approach only requires noisy LiDAR data and can implicitly model the data-dependent uncertainty of point measurability.
As we discuss in Section~\ref{sec:reconstruction}, our learned generator can be used to estimate realistic point-drops from given observable points.

\subsection{Modeling Image Noises}

Some studies leveraged noisy training data for denoising images~\cite{pmlr-v80-lehtinen18a} and synthesizing novel clean images~\cite{kaneko2020noise,bora2018ambientgan,li2018learning}.
Lehtinen \textit{et al.}~\cite{pmlr-v80-lehtinen18a} eliminated various types of synthetic noises, including multiplicative Bernoulli noise, which randomly masked a pixel to zero with a certain probability.
Bora \textit{et al.}~\cite{bora2018ambientgan} learned GANs from partially-zeroed images.
We assume that point-drops in LiDAR sensing can be categorized as this type of noise.
However, these studies~\cite{pmlr-v80-lehtinen18a,bora2018ambientgan} assume a predefined noise model, such as pixel-level or patch-level dropouts with a fixed probability.
Kaneko and Harada~\cite{kaneko2020noise} proposed GAN architectures that can estimate noise distribution, but multiplicative binary noises are not covered.
Li \textit{et al.}~\cite{li2018learning} learn the distribution of binary noises, however, the noise is independent of signals and approximated with sigmoid outputs.

As in \cite{pmlr-v80-lehtinen18a,bora2018ambientgan,li2018learning}, we assume multiplicative Bernoulli noises to mimic the point-drops on the LiDAR scan map.
However, we do not set the probability; instead, we aim to learn the pixel-wise probability model.
Specifically, in our case, the noise distribution type is defined, but the noise level and the generative relationship to the depth modality are unknown owing to the complex physical factors in measurement.
Meanwhile, generating binary masks with Bernoulli distribution is not differentiable, which means inability to learn the parameterized probability model by backpropagation.
Therefore, this study employs the Gumbel-Softmax trick~\cite{Jang2017categorical,maddison2017concrete}, a reparametrization approach for discrete sampling.

\section{Our Approach}

\subsection{Data Representation}
\label{sec:data}

This paper assumes a bijective image representation~\cite{caccia2019deep}, which is directly acquired from horizontal scanning of multiple laser receptors aligned vertically.
For example, a LiDAR that emits $W$ pulses for $H$ elevation angles produces $H\times W$ points with measured distances $x$, which can be considered as a cylindrical depth map with a size of $H\times W$.
An existing study~\cite{caccia2019deep} trained GANs with the depth map representation.
However, we found it difficult to learn stably without their preprocessing that narrows the spatial range of the depth map.
Instead, we propose another approach for data representation motivated by monocular depth estimation.
In a typical setting of LiDARs, most pixels of the acquired depth map represent a near-to-mid range, and most of the dynamic range is occupied by few distant points in a long-tail distribution.
Therefore, to gain the dynamic range of the majority regions, we transform the LiDAR depth map into an inverse form $1/x$, \textit{i.e.}, an inverse depth map.
We will discuss how we can synthesize the cylindrical inverse depth maps throughout this paper.

\subsection{Decomposed Synthesis of LiDAR Scans}
\label{sec:approach:decomposed}

A GAN typically consists of two networks: a generator $G_\theta$ and a discriminator $D_\phi$, where $\theta$ and $\phi$ are trainable parameters.
In image synthesis tasks, $G_\theta$ maps a latent variable $\bm{z}\sim N(0,\bm{I})$ to an image $\bm{x}_{\bm{z}}=G_\theta(\bm{z})$, whereas $D_\phi$ distinguishes the generated image $\bm{x}_{\bm{z}}$ from sampled real images $\bm{x}_\mathrm{real}$.
The networks are trained in an alternating fashion by minimizing the adversarial objective, \textit{e.g.}, the following non-saturating loss~\cite{goodfellow2014generative}:
\begin{eqnarray}
	\label{eq:adversarial_d}
	\mathcal{L}_D=-\mathbb{E}_x[\log D_\phi(\bm{x}_\mathrm{real})]-\mathbb{E}_z[\log (1-D_\phi(G_\theta(\bm{z})))] \\
	\label{eq:adversarial_g}
	\mathcal{L}_G=-\mathbb{E}_z[\log D_\phi(G_\theta(\bm{z}))].
\end{eqnarray}

Here, $G_\theta$ is modeled as a decoder convolutional network that gradually upscales the spatial resolution toward the final image.
Although synthesizing natural images~\cite{karras2020analyzing} has been successfully achieved, we found it difficult to learn noisy depth maps with a stack of na\"{\i}ve convolutions.

Let us consider modeling a scene where a vehicle is moving, for example.
If all the points are observable on the LiDAR receptor,  the shape of the vehicle would be smoothly morphed on the depth map.
In this case, generating a sequence of the depth maps is relatively easy if we move a sampling point in the continuous data space by manipulating a set of convolution kernels.
However, in practice, some points are randomly dropped at the pixel level because of complex physical factors (such as mirror diffusion and material reflectance).
In these cases, the scattered point-drops make the morphing discrete and cause learning difficulty.
Our key idea is to simulate the point-drop phenomenon by a probabilistically invertible function to learn the underlying manifold of smooth depth separately.

To this end, we introduce a decomposed depth map representation into adversarial training.
An overview of our approach is depicted in Fig.~\ref{fig:model}.
First, we assume a dense inverse depth map $\tilde{\bm{x}}_{\bm{z}}\in\mathbb{R}^{H\times W}$ and a measurability map $\bm{\pi}_{\bm{z}}\in\mathbb{R}^{H\times W}$ that represents the probability of laser reflection (\textit{e.g.}, mirror-like materials have low confidence in returning lasers due to the diffuse reflection, which results in point-drops).
We sample a binary mask $\bm{m}_{\bm{z}}\in\{0,1\}^{H\times W}$ from the measurability map $\bm{\pi}_{\bm{z}}$.
\begin{equation}
	\label{eq:bernoulli}
	\bm{m}_{\bm{z}} \sim \mathrm{Bernoulli}\left(\bm{\pi}_{\bm{z}}\right)
\end{equation}

The final synthetic scan $\bm{x}_{\bm{z}}$ is produced by masking the dense inverse depth $\tilde{\bm{x}}_{\bm{z}}$ using the binary mask $\bm{m}_{\bm{z}}$:
\begin{equation}
	\label{eq:final_depth}
	\bm{x}_{\bm{z}} = \bm{m}_{\bm{z}}\odot\tilde{\bm{x}}_{\bm{z}} + \left(1-\bm{m}_{\bm{z}}\right)\odot\alpha,
\end{equation}
where $\odot$ is an element-wise product, and $\alpha$ is a constant value representing the point-drop.

Note that sampling $\bm{m}_{\bm{z}}$ is not differentiable, and the gradients cannot be propagated downstream.
Therefore, we reparameterize $\bm{m}_{\bm{z}}$ with the straight-through (ST) Gumbel-Sigmoid\footnote{Gumbel-Sigmoid is a binary case of Gumbel-Softmax.} distribution~\cite{Jang2017categorical,maddison2017concrete} to estimate the gradients.
We first introduce the continuous relaxation $\tilde{\bm{m}}_{\bm{z}}$ as follows:
\begin{equation}
	\label{eq:gumbel_sigmoid:soft}
	\tilde{\bm{m}}_{\bm{z}} = \mathrm{sigmoid}\left(\frac{\bm{e}+\bm{g}_1-\bm{g}_2}{\tau}\right), \bm{e} = \ln{\left(\frac{\bm{\pi}_{\bm{z}}}{1-\bm{\pi}_{\bm{z}}}\right)},
\end{equation}
where $\bm{e}$ is a logit of $\bm{\pi}_{\bm{z}}$ and $\bm{g}_1,\bm{g}_2\in\mathbb{R}^{H\times W}$ are i.i.d. samples from $ \mathrm{Gumbel}(0,\bm{I})$, which perturbs $\bm{e}$ at the pixel level. Moreover,
$\tau$ is a hyperparameter called temperature, which controls the slope of the sigmoid function.
With a low value of $\tau$, the soft mask $\tilde{\bm{m}}_{\bm{z}}$ approaches a binary mask, but the variance of the gradients is large.
Finally, the soft mask $\tilde{\bm{m}}_{\bm{z}}$ is discretized into the binary mask $\bm{m}_{\bm{z}}$ at each pixel location $(i,j)$ as follows:
\begin{equation}
	\label{eq:threshold}
	\bm{m}_{\bm{z}}^{i,j}=\begin{cases}
	1 & \tilde{\bm{m}}_{\bm{z}}^{i,j}\geq0.5 \\
	0 & \tilde{\bm{m}}_{\bm{z}}^{i,j}<0.5
	\end{cases}
\end{equation}

ST Gumbel-Sigmoid~\cite{Jang2017categorical,maddison2017concrete} approximates this thresholding by an identity function to enable gradient propagation:
\textit{i.e.}, we use the stochastic binary mask $\bm{m}_{\bm{z}}$ in the forward step, and we approximate the gradients by $\tilde{\bm{m}}_{\bm{z}}$ during the backward step.
In summary, instead of modeling raw data directly, our generator $G_\theta$ aims to jointly produce the dense inverse depth $\tilde{\bm{x}}_{\bm{z}}$ and the associated confidence of point measurability $\bm{\pi}_{\bm{z}}$.
Discriminator $D_\phi$ only determines the realness of the processed inverse depth map $\bm{x}_{\bm{z}}$ in Eq.~\ref{eq:final_depth}.

\subsection{Multilevel Gumbel Sampling}
\label{sec:approach:multi}

As introduced above, the pixel-level Gumbel sampling offers the differentiable binarizer.
The stochastic behavior can learn the pixel-level uncertainty.
However, we observed that another level of uncertainty appears in real LiDAR data.
For example, on the depth map of KITTI, some cars have point-drops on only the transparent windows, while others are missing most of the body.
We assume colors and materials cause this object-level multimodality of point drops even for the same shape.
The pixel-level Gumbel sampling is hard to catch this distribution, and as a result, the variation would be leaked into the latent space.
We introduce a simple extension to separate the object-level uncertainty from the latent space.

\begin{figure}[tb]
	\centering
	\includegraphics[width=0.9\hsize]{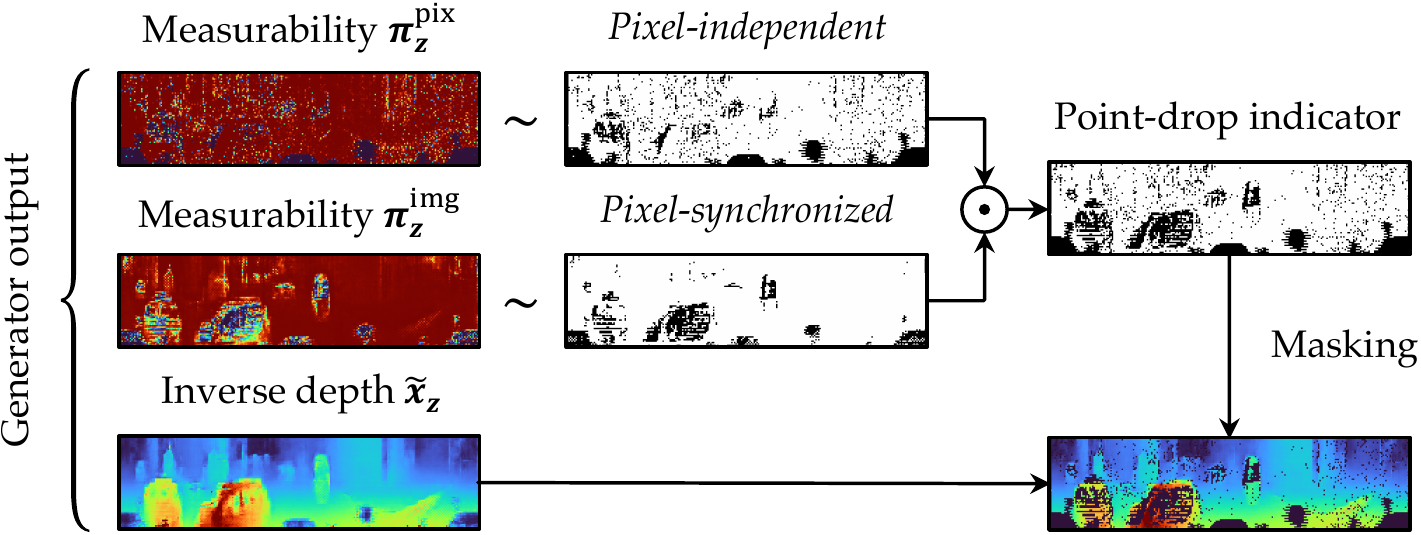}
	\caption{Multilevel Gumbel sampling. $\odot$ denotes an element-wise product.}
	\label{fig:model:multilevel}
\end{figure}

Revisiting Eq.~\ref{eq:gumbel_sigmoid:soft}, we made \textit{pixel-independent} perturbations to model the measurability.
Here, we call the measurability and the sampled mask as $\bm{\pi}_{\bm{z}}^\mathrm{pix}$ and $\bm{m}_{\bm{z}}^\mathrm{pix}$, respectively.
We now consider another measurability branch $\bm{\pi}_{\bm{z}}^\mathrm{img}$ to be added \textit{pixel-synchronized} perturbations, where we sample scalars $g_1, g_2\sim\mathrm{Gumbel}(0,1)$ in training and cancel them in testing to switch to a deterministic mask $\bm{m}_{\bm{z}}^\mathrm{img}$.
The final mask $\bm{m}_{\bm{z}}$ is determined by element-wise product $\bm{m}_{\bm{z}}^\mathrm{pix}\odot\bm{m}_{\bm{z}}^\mathrm{img}$.
Fig.~\ref{fig:model:multilevel} shows a pipeline of the multilevel Gumbel sampling.

\section{Synthesis by Latent Sampling}

We first report the synthesis performance of our proposed methods by evaluating the similarity between the distributions of real data and sampled synthetic data.
We further discuss the reconstruction performance in Section~\ref{sec:reconstruction}.

\subsection{Dataset}

We built the inverse depth maps from two types of LiDAR datasets: KITTI~\cite{geiger2013vision} and MPO~\cite{martinez2019fukuoka}, which have different vertical resolutions and scenes.
The LiDARs of both the KITTI and MPO datasets have the same distance range of $(0.9 m,120 m)$, where we take the reciprocal of the distances and normalize the range into $(-1,1)$ for the training stability of GANs.
When reconstructing point clouds from the inverse depth map, we use the per-pixel average angles calculated from the training set.

\subsubsection{KITTI}

We used the KITTI odometry~\cite{geiger2013vision} dataset, which is a benchmark for odometry evaluation.
The dataset provides 22 trajectories of LiDAR scans measured by the Velodyne HDL-64E\footnote{\url{https://velodynelidar.com/products/hdl-64e/}} scanner, where each scan has 64 layers vertically.
It includes 19,310 scans for training, 4,071 scans for validation, and 20,351 scans for testing.
The provided data are processed sequences of Cartesian coordinate points ordered by azimuth and elevation angles.
Most studies applied spherical projection to create a depth map from the point sets. However, this approach produces sampling artifacts due to the scanner's nonlinear vertical spacing\footnote{The HDL-64E scanner has several versions with different vertical spacing, but it was not unified in studies with KITTI.}.
As in Caccia~\textit{et al.}~\cite{caccia2019deep}, we first chunk the ordered sequence into 64 sub-sequences, where each represents one elevation angle.
We then subsample 256 points for each sub-sequence and stack them to form a $64\times256$ inverse depth map.
Unlike~\cite{caccia2019deep}, we do not clip the vertical angle and the distance range and do not fill in the point-drops with adjacent pixels\footnote{\url{https://github.com/pclucas14/lidar_generation}}.

\subsubsection{MPO}

We also use the Multimodal Panoramic 3D Outdoor dataset (MPO)~\cite{martinez2019fukuoka}.
In particular, we use a ``sparse'' dataset of MPO, which includes a total of 34,200 LiDAR scans from 60 trajectories.
The dataset has diversity in geometry and point-drop distribution because it was constructed to classify six different outdoor scenes: coast, forest, indoor parking, outdoor parking, residential area, and urban area.
Each scan is a sequence of Cartesian coordinate points with vertical angle IDs, obtained using Velodyne HDL-32E\footnote{\url{https://velodynelidar.com/products/hdl-32e/}} scanner.
In this study, we split the trajectories into 20,322 scans for training, 3,787 scans for validation, and 10,091 scans for testing.
We chunk the sequence according to the vertical angle ID and create a $32\times256$ inverse depth map similar to the KITTI procedure.

\subsection{Models}

\subsubsection{Baseline}

\begin{table}[tb]
	\centering
	\scriptsize
	\caption{Baseline architecture. $H$ and $W$ are the height and width of the inverse depth map, respectively. $\dagger$The shape is replaced with $2\times H\times W$ for DUSty-I and $3\times H\times W$ for DUSty-II.}
	\label{tab:baseline_arch}
	\begin{tabularx}{\hsize}{r@{\hspace{2mm}}l@{\hspace{2mm}}C@{\hspace{1mm}}c@{\hspace{1mm}}C@{\hspace{3mm}}c@{\hspace{3mm}}c@{\hspace{1mm}}c@{\hspace{1mm}}C@{\hspace{1mm}}c@{\hspace{1mm}}C}
		\toprule
		& Layer                         & \multicolumn{3}{l}{Kernel size} & Nonlinearity & \multicolumn{5}{c}{Output shape}                                                                       \\
		\midrule
		\multirow{6}{*}{\rotatebox[origin=c]{90}{Generator $G_{\theta}$}}   & (Input)                               &        & --       &         & --         & $512$ & $\times$ & $1$    & $\times$ & $1$           \\
		                                                                    & Transposed Conv.                      & $H/16$ & $\times$ & $ W/16$ & Leaky ReLU & $512$ & $\times$ & $H/16$ & $\times$ & $W/16$        \\
		                                                                    & Transposed Conv.                      & $4$    & $\times$ & $4$     & Leaky ReLU & $256$ & $\times$ & $H/8$  & $\times$ & $W/8$         \\
		                                                                    & Transposed Conv.                      & $4$    & $\times$ & $4$     & Leaky ReLU & $128$ & $\times$ & $H/4$  & $\times$ & $W/4$         \\
		                                                                    & Transposed Conv.                      & $4$    & $\times$ & $4$     & Leaky ReLU & $64$  & $\times$ & $H/2$  & $\times$ & $W/2$         \\
		                                                                    & Transposed Conv.                      & $4$    & $\times$ & $4$     & Tanh       & $1$   & $\times$ & $H$    & $\times$ & $W^{\dagger}$ \\
		\midrule
		\multirow{7}{*}{\rotatebox[origin=c]{90}{Discriminator $D_{\phi}$}} & (Input)                               &        & --       &         & --         & $1$   & $\times$ & $H$    & $\times$ & $W$           \\
		                                                                    & Blur filtering~\cite{kaneko2020noise} & $3$    & $\times$ & $3$     & --         & $2$   & $\times$ & $H$    & $\times$ & $W$           \\
		                                                                    & Convolution                           & $4$    & $\times$ & $4$     & Leaky ReLU & $64$  & $\times$ & $H/2$  & $\times$ & $W/2$         \\
		                                                                    & Convolution                           & $4$    & $\times$ & $4$     & Leaky ReLU & $128$ & $\times$ & $H/4$  & $\times$ & $W/4$         \\
		                                                                    & Convolution                           & $4$    & $\times$ & $4$     & Leaky ReLU & $256$ & $\times$ & $H/8$  & $\times$ & $W/8$         \\
		                                                                    & Convolution                           & $4$    & $\times$ & $4$     & Leaky ReLU & $512$ & $\times$ & $H/16$ & $\times$ & $W/16$        \\
		                                                                    & Convolution                           & $H/16$ & $\times$ & $ W/16$ & --         & $1$   & $\times$ & $1$    & $\times$ & $1$           \\
		\bottomrule
	\end{tabularx}
\end{table}

Table~\ref{tab:baseline_arch} shows the network architectures of our baseline GAN.
The architecture design is inspired by the existing work~\cite{caccia2019deep}. However, we modify the output shape for our task, remove all normalization layers, and only use the Leaky ReLU nonlinearity with a negative slope of 0.2.
We add vertical/horizontal blur filtering~\cite{kaneko2020noise} as the first layer of the discriminator, to mitigate the learning difficulty of a discrete distribution.
As in existing studies~\cite{schubert2019circular,nakashima2018learning}, we replace the zero-padding of all convolutional layers with horizontal circular padding, where the right and left boundaries are padded with the pixels on the opposite sides.
It enables the model to spread the receptive fields out of the boundaries so that it can process the cylindrical tensors.

\subsubsection{DUSty-I (ours)}

For a fair comparison, in this method, we only modify the last convolutional layer of the baseline generator.
The last layer produces 2-channel outputs for inverse depth and measurability.
The tanh nonlinearity is applied to only the inverse depth output.
The measurability output is transformed into a binary mask with pixel-level Gumbel sampling, as introduced in Section~\ref{sec:approach:decomposed}.

\subsubsection{DUSty-II (ours)}

In this method, we add an extra channel to the measurability output of the DUSty-I model to decompose the Gumbel sampling into pixel-level and image-level, as introduced in Section~\ref{sec:approach:multi}.
The same temperature $\tau$ is used for both pixel-level and image-level Gumbel sampling.

\subsection{Implementation Details}

We set the maximum distance to the point-drop constant $\alpha$ in Eq.~\ref{eq:final_depth}.
Moreover, we observed that a low value of temperature $\tau$ in Eq.~\ref{eq:gumbel_sigmoid:soft} slowed the convergence despite a better approximation of the binary masks.
This study uses a fixed $\tau=1$ in all experiments.
For Gumbel sampling from measurability $\bm{\pi}$ in Eq.~\ref{eq:gumbel_sigmoid:soft}, our network directly outputs the logit $\bm{e}$, instead of $\bm{\pi}=\mathrm{sigmoid}(\bm{e})$.
For adversarial training, we use the non-saturating loss in Eq.~\ref{eq:adversarial_d}~and~\ref{eq:adversarial_g}, with a $R_1$ gradient penalty~\cite{karras2020analyzing}. The penalty coefficient was set to 1.
All parameters were updated by Adam~\cite{DBLP:journals/corr/KingmaB14} optimizer for 25M iterations with a learning rate of 0.002 and a batch size of 32.
We apply the equalized learning rate~\cite{karras2018progressive} for all trainable layers, where the parameters are initialized with $\mathcal{N}(0,1)$ and scaled by He's initialization constant at runtime.
Moreover, we apply DiffAugment by Zhao \textit{et al.}~\cite{zhao2020diffaugment}, which composed of \textit{color}, \textit{translation}, and \textit{cutout} augmentations to the discriminator inputs.
For the \textit{translation} augmentation, we circulate the inputs horizontally.
DiffAugment was not required for training stability but greatly improved the quality of inverse depth maps.
We take the exponential moving average for generator parameters $\theta$.
We implemented our networks in PyTorch and performed distributed training on two NVIDIA Titan RTX GPUs.
The training required approximately 22 hours for each model.
The code will be available at \url{https://github.com/kazuto1011/dusty-gan}.

\subsection{Evaluation Metrics}

We measured four types of distributional similarities between the sets of reference and generated point clouds~\cite{yang2019pointflow}: Jensen--Shannon divergence (JSD) for quality, coverage (COV) for diversity, minimum matching distance (MMD) for quality, and 1-nearest neighbor accuracy (1-NNA) for both quality and diversity evaluation.
We found that the synthesis evaluation on LiDAR point clouds had an extremely high cost, particularly for calculating the distance matrix for the sets of point clouds in COV, MMD, and 1-NNA.
For efficiency, we first subsampled the series of test data to be 5,000 data in total, and for each, we also randomly subsampled 2,048 points from the full $H\times W$ points by farthest point sampling.
We generated 5,000 data from each model and reduced the number of points in the same manner.
We used the Chamfer distance to measure the pairwise similarity for the distance matrix of point clouds.
Additionally, we computed the sliced Wasserstein distance (SWD)~\cite{karras2018progressive} to measure the patch-based image similarity for the inverse depth maps.
For all metrics, we report the mean scores with the standard deviation over five runs with different latent codes.

\subsection{Quantitative Results}
\label{sec:synthesis:quantitative}

\begin{figure}[tb]
	\footnotesize
	\centering
	\begin{minipage}[c]{\hsize}
		\centering
		\includegraphics[width=\hsize]{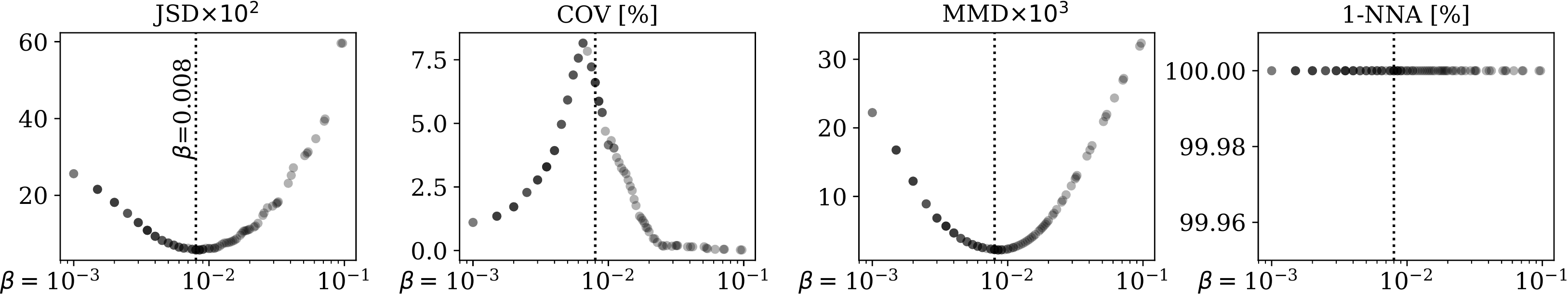}
		\\
		(a) Scores on KITTI validation set
	\end{minipage}
	\\[1mm]
	\begin{minipage}[c]{\hsize}
		\centering
		\includegraphics[width=\hsize]{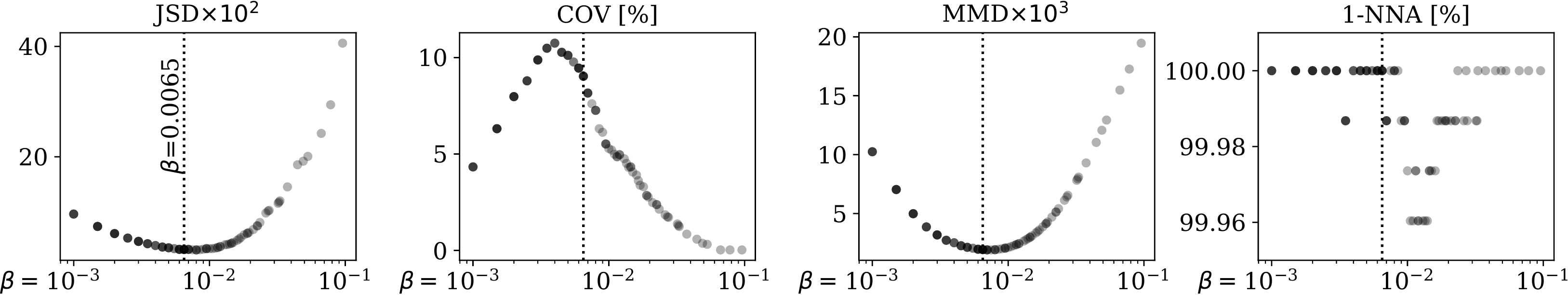}
		\\
		(b) Scores on MPO validation set
	\end{minipage}
	\caption{Impact of relative tolerance $\beta$ for the baseline method. We optimized $\beta$ based on the weighted score and set $0.008$ for KITTI and $0.0065$ for MPO (dotted vertical lines).}
	\label{fig:tol_tuning}
\end{figure}

\begin{table*}[t]
	\centering
	\caption{Quantitative comparison of synthesis performance. $\downarrow$: the lower the better. $\uparrow$: the higher the better.}
	\label{tab:results_all}
	\begin{tabularx}{\hsize}{llCCCcC}
		\toprule
		\multirow{2}{*}{Dataset} & \multirow{2}{*}{Method}  & JSD $\times10^2$ $\downarrow$ & COV [\%] $\uparrow$   & MMD $\times10^3$ $\downarrow$ & 1-NNA [\%] $\downarrow$ & SWD $\downarrow$       \\
		                         &                          & (3D quality)                  & (3D diversity)        & (3D quality)                  & (3D quality/diversity)  & (2D quality)           \\
		\midrule
		\multirow{4}{*}{KITTI}   & Baseline                 & $6.45 \pm 0.06$               & $4.99 \pm 0.04$       & $2.36 \pm 0.03$               & $99.99 \pm 0.00$        & $0.158 \pm 0.011$      \\
		                         & DUSty-I (\textbf{ours})  & $\bm{2.85} \pm 0.01$          & $\bm{38.08} \pm 0.55$ & $1.14 \pm 0.01$               & $\bm{93.69} \pm 0.19$   & $0.167 \pm 0.010$      \\
		                         & DUSty-II (\textbf{ours}) & $3.54 \pm 0.07$               & $38.05 \pm 0.65$      & $\bm{1.12} \pm 0.01$          & $94.62 \pm 0.30$        & $\bm{0.151} \pm 0.011$ \\
		\cmidrule{2-7}
		                         & Training set             & 0.93                          & 35.02                 & 0.87                          & 96.72                   & 0.182                  \\
		\midrule
		\multirow{4}{*}{MPO}     & Baseline                 & $2.53 \pm 0.03$               & $6.10 \pm 0.12$       & $2.03 \pm 0.01$               & $99.31 \pm 0.03$        & $0.180 \pm 0.013$      \\
		                         & DUSty-I (\textbf{ours})  & $\bm{1.47} \pm 0.02$          & $22.85 \pm 0.34$      & $\bm{1.53} \pm 0.01$          & $95.03 \pm 0.16$        & $0.174 \pm 0.014$      \\
		                         & DUSty-II (\textbf{ours}) & $1.71 \pm 0.02$               & $\bm{30.94} \pm 0.34$ & $1.54 \pm 0.01$               & $\bm{94.84} \pm 0.19$   & $\bm{0.160} \pm 0.018$ \\
		\cmidrule{2-7}
		                         & Training set             & 0.74                          & 34.88                 & 1.52                          & 87.09                   & 0.158                  \\
		\bottomrule
	\end{tabularx}
\end{table*}

It is non-trivial for the baseline model to produce values that exactly match the point-drop constant $\alpha$.
Therefore, we first optimize a relative tolerance $\beta$ to decide if an interest pixel $\bm{x}_{\bm{z}}^{i,j}$ is $\alpha$ based on
\begin{equation}
	\frac{\lVert\bm{x}_{\bm{z}}^{i,j}-\alpha\rVert}{x_{\mathrm{max}}-x_{\mathrm{min}}}\leq\beta,
\end{equation}
where $x_{\mathrm{max}}-x_{\mathrm{min}}$ is a range of the inverse depth.
$\beta$ is optimized by HyperOpt~\cite{bergstra2013making} within $[10^{-3},10^{-1}]$ for 100 steps to minimize the following weighted 3D score, 
$\mathrm{JSD} \times 10 - \mathrm{COV} + \mathrm{MMD} \times 10^2 + \mathrm{1{\text -}NNA}$, on the validation set for each dataset.
For efficiency, we reduce the number of points to 512.
Fig.~\ref{fig:tol_tuning} shows trade-off results because the point-drop tolerance erodes the available range of inverse depth.
We employ $\beta=0.008$ for KITTI and $\beta=0.0065$ for MPO.
We note that our noise-aware model DUSty \textit{does not} require the tolerance, \textit{i.e.}, $\beta=0$.

In Table~\ref{tab:results_all}, we compare the synthesis performance among the baseline and our methods.
On both KITTI~\cite{geiger2013vision} and MPO~\cite{martinez2019fukuoka} datasets, our methods outperformed the baseline on all the 3D metrics.
On the SWD score, the baseline was better than DUSty-I on KITTI, while our methods outperformed in the other cases.
We consider that our sampling-based binary mask still has an ``appearance'' gap with the truth distribution of point-drops compared with the direct modeling by the baseline.
Meanwhile, the 3D results on four metrics indicate that our decomposition approach has a positive effect on the disentangled modeling of inverse depth and point-drop.

\subsection{Qualitative Results}

\begin{figure}[t]
	\centering
	\includegraphics[width=\hsize]{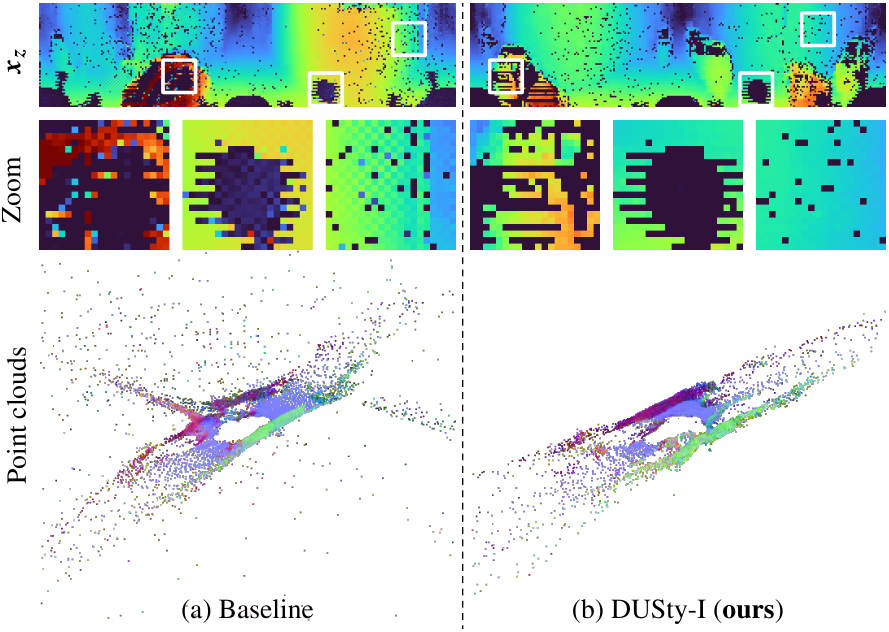}
	\caption{Qualitative comparison of inverse depth outputs $\bm{x}_{\bm{z}}$, zoomed-in regions (white boxes in $\bm{x}_{\bm{z}}$), and points clouds, with and without the proposed measurability learning. Best viewed in color.}
	\label{fig:zoomed}
\end{figure}

\begin{figure}[t]
	\centering
	\includegraphics[width=\hsize]{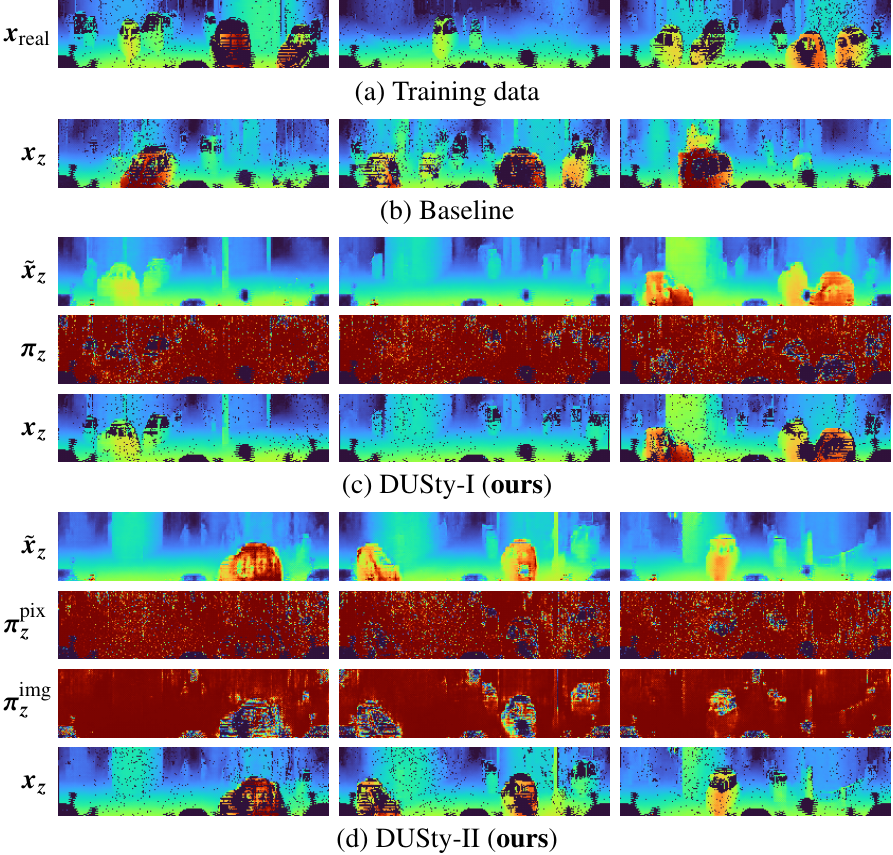}
	\caption{Qualitative comparison on the KITTI~\cite{geiger2013vision} dataset. Each column is from a unique latent code. $\bm{x}_{\mathrm{real}}$ and $\bm{x}_{\bm{z}}$ denotes real LiDAR data and generated LiDAR data, respectively. $\tilde{\bm{x}}_{\bm{z}}$ and $\bm{\pi}_{\bm{z}}$ denotes the inverse depth and measurability outputs from our proposed models, respectively.}
	\label{fig:examples:kitti}
\end{figure}

\begin{figure}[t]
	\centering
	\includegraphics[width=\hsize]{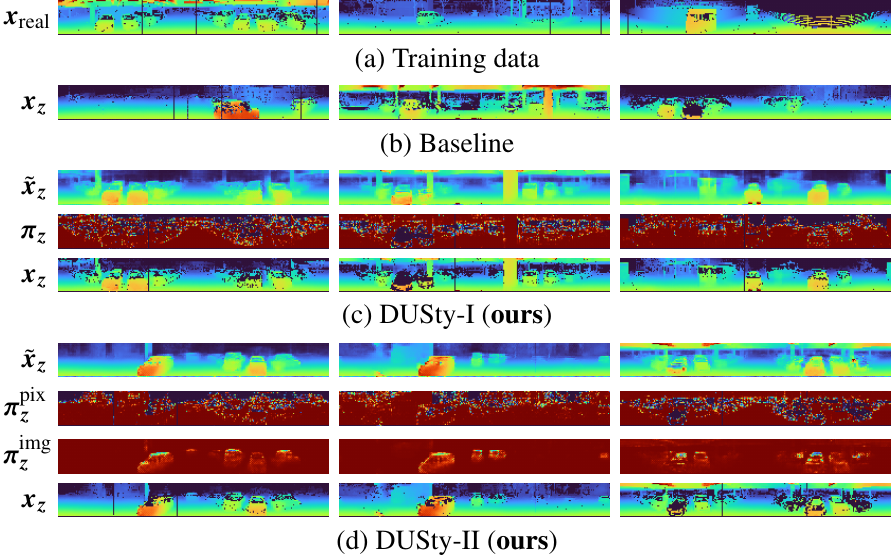}
	\caption{Qualitative comparison on the MPO~\cite{martinez2019fukuoka} dataset. Each column is from a unique latent code.}
	\label{fig:examples:mpo}
\end{figure}

Fig.~\ref{fig:zoomed} compares inverse depth maps generated from the baseline and our DUSty-I model on the KITTI dataset.
Our method succeeds in representing the sharp jump edges around point-drops. However, the baseline has interpolated pixels that can be unexpected noise in point clouds.
Moreover, we show some synthetic examples of all methods for KITTI in Fig.~\ref{fig:examples:kitti} and MPO in Fig.~\ref{fig:examples:mpo}.
We can see that our methods DUSty-I and DUSty-II successfully learned a complete inverse depth maps in $\tilde{\bm{x}}_{\bm{z}}$ and drop uncertainty in $\bm{\pi}_{\bm{z}}$.
Regarding $\bm{\pi}_{\bm{z}}$, the results show that the sampling variation would occur on the boundaries of vehicle-like objects and the scattered noises.
In contrast, the steady point-drops are represented as sufficiently low measurability, such as the ego-vehicle shadows in KITTI and the sky regions in MPO.
Furthermore, DUSty-II separately learned the pixel correlation of the drop uncertainty in $\bm{\pi}_{\bm{z}}^{\mathrm{img}}$.

\section{Reconstruction by Latent Space Exploration}
\label{sec:reconstruction}

With trained GANs, we can reproduce the given data by optimizing a latent code~\cite{karras2020analyzing,menon2020pulse}, where the task is called \textit{GAN inversion}.
In this section, we optimize the latent code $\bm{z}$ to reconstruct unseen LiDAR data from KITTI and MPO test sets.
We demonstrate our decomposition approach is easier to find a matching latent code.

\subsection{Objective}

To reproduce a given inverse depth $\bm{x}_{\mathrm{target}}$, we optimize the latent code $\bm{z}$ by minimizing the L1 distance between $\bm{x}_{\mathrm{target}}$ and generated inverse depth $\tilde{\bm{x}}_{\bm{z}}$ before masking:
\begin{equation}
	\hat{\bm{z}}=\arg \min_{\bm{z}} \frac{\sum_{i,j} \mathds{1}[\bm{x}_{\mathrm{target}}\neq\alpha]^{i,j} \lVert \bm{x}_{\mathrm{target}}^{i,j}-\tilde{\bm{x}}_{\bm{z}}^{i,j} \rVert_{1}}{\sum_{i,j} \mathds{1}[\bm{x}_{\mathrm{target}}\neq\alpha]^{i,j}},
\end{equation}
where $\mathds{1}[\bm{x}_{\mathrm{target}}\neq\alpha]\in\{0,1\}^{H\times W}$ is a point-drop indicator of the target data $\bm{x}_{\mathrm{target}}$, and $(i,j)$ is the 2D location of the inverse depth map.
The latent $\bm{z}\in\mathbb{R}^d$ is updated for 1,000 iterations by the Adam optimizer with a learning rate of $0.1$.
Considering the nature of high-dimensional Gaussian priors, the search space of the latent $\bm{z}$ is constrained to a surface of a hypersphere with a radius of $\sqrt{d}$~\cite{menon2020pulse}.
To avoid sticking in local minima, we add a Gaussian noise $\epsilon\sim\mathcal{N}(0,0.05t^2\bm{I})$, where $t$ goes from 1 to 0 in iterations~\cite{karras2020analyzing}.
The optimization required approximately 4 seconds for each sample.

\subsection{Quantitative Results}

To assess the reconstruction performance, we compute standard metrics in depth estimation~\cite{geiger2013vision}: relative absolute error (Abs Rel), relative squared error (Sq Rel), root mean squared error (RMSE), RMSE on logarithmic depth (RMSE log), and the ratio of pixels with relative error $\delta$ under the threshold ($\delta<1.25^i, i=1,2,3$).
Moreover, we also compute the Chamfer distance (CD) between point clouds from $\bm{x}_{\mathrm{target}}$ and $\bm{x}_{\hat{\bm{z}}}$.
Table~\ref{tab:reconstruction:quantitative} shows the results of the baseline and our proposed methods.
Our method outperformed the baseline on all the metrics, and DUSty-I was slightly better than DUSty-II in most cases.

\begin{table*}[t]
	\centering
	\caption{Quantitative comparison of reconstruction performance. $\downarrow$: the lower the better. $\uparrow$: the higher the better.}
	\label{tab:reconstruction:quantitative}
	\resizebox*{\hsize}{!}{
		\begin{tabularx}{\hsize}{llCCCCCCCC}
			\toprule
			&                          & 3D error $\downarrow$ & \multicolumn{4}{c}{2D error $\downarrow$} & \multicolumn{3}{c}{2D accuracy [\%] $\uparrow$}                                                              \\
			\cmidrule(lr){3-3} \cmidrule(lr){4-7} \cmidrule(lr){8-10}
			Dataset                & Method                   & CD $\times10^4$ & Abs Rel        & Sq Rel         & RMSE           & RMSE log       & $\delta<1.25$  & $\delta<1.25^2$ & $\delta<1.25^3$ \\
			\midrule
			\multirow{3}{*}{KITTI} & Baseline                 & 5.31            & 0.199          & 12.394         & 9.667          & 0.280          & 89.11          & 95.21           & 97.51           \\
			                       & DUSty-I (\textbf{ours})  & \textbf{1.64}   & \textbf{0.074} & 0.555          & 3.722          & \textbf{0.155} & \textbf{92.09} & \textbf{96.94}  & \textbf{98.69}  \\
			                       & DUSty-II (\textbf{ours}) & 1.66            & 0.077          & \textbf{0.546} & \textbf{3.663} & 0.157          & 91.91          & 96.93           & \textbf{98.69}  \\
			\midrule
			\multirow{3}{*}{MPO}   & Baseline                 & 9.33            & 0.197          & 9.008          & 10.523         & 0.308          & 84.15          & 92.26           & 95.61           \\
			                       & DUSty-I (\textbf{ours})  & \textbf{1.25}   & \textbf{0.109} & \textbf{0.878} & \textbf{4.930} & \textbf{0.229} & \textbf{85.90} & \textbf{94.02}  & \textbf{97.11}  \\
			                       & DUSty-II (\textbf{ours}) & 1.28            & 0.111          & 1.013          & 5.079          & 0.232          & 85.83          & 93.89           & 97.00           \\
			\bottomrule
		\end{tabularx}
	}
\end{table*}

\subsection{Qualitative Results}

\begin{figure}[t]
	\centering
	\includegraphics[width=\hsize]{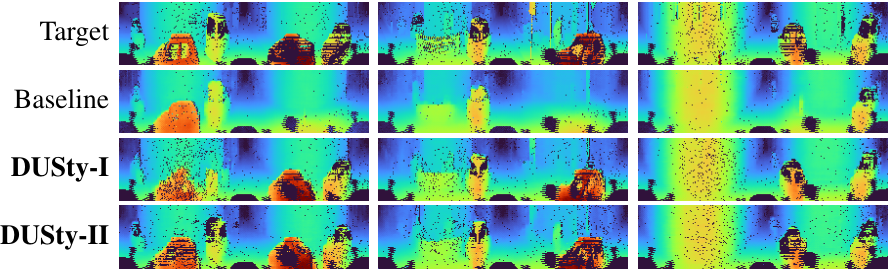}
	\caption{Qualitative comparison of the reconstructed inverse depth maps. Our method is better at reproducing the geometric features of the given targets and generating realistic point-drops.}
	\label{fig:compare_reconstruction}
\end{figure}

Fig.~\ref{fig:compare_reconstruction} compares the reconstructed inverse depth maps $\bm{x}_{\hat{\bm{z}}}$ from the baseline and our methods.
The baseline failed to reconstruct some foreground objects ($\hat{\bm{z}}\leftarrow\bm{x}_{\mathrm{target}}$), contrary to the synthesis results in Fig.~\ref{fig:zoomed}~and~\ref{fig:examples:kitti} ($\bm{z}\rightarrow\bm{x}_{\bm{z}}$).
In contrast, our method reproduced the details and also synthesized more realistic point-drops.
This indicates that our decomposition trick yields a better embedding of the geometric styles of LiDAR scans.

\subsection{Applications}

\begin{figure*}[t]
	\centering
	\includegraphics[width=0.49\hsize]{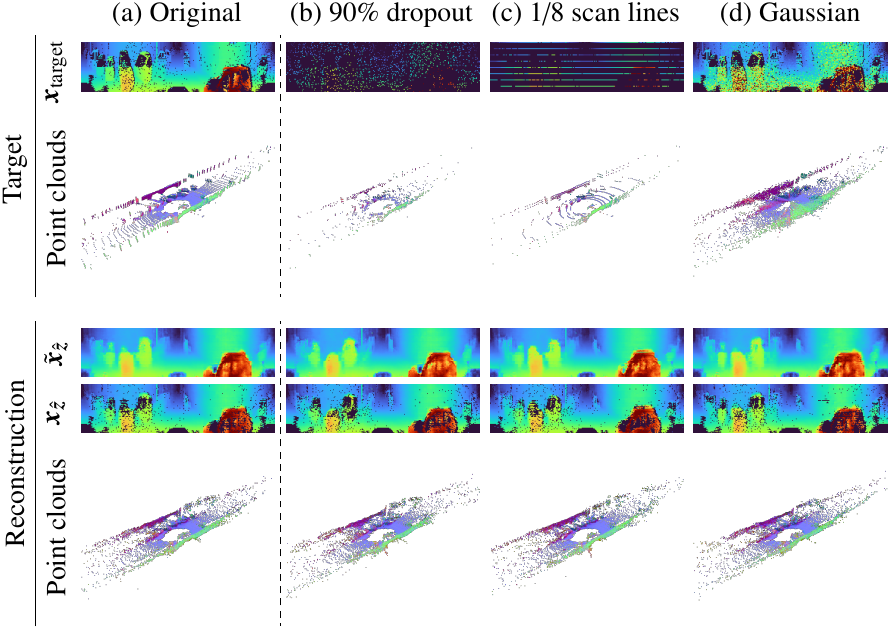}
	\hfill
	\includegraphics[width=0.49\hsize]{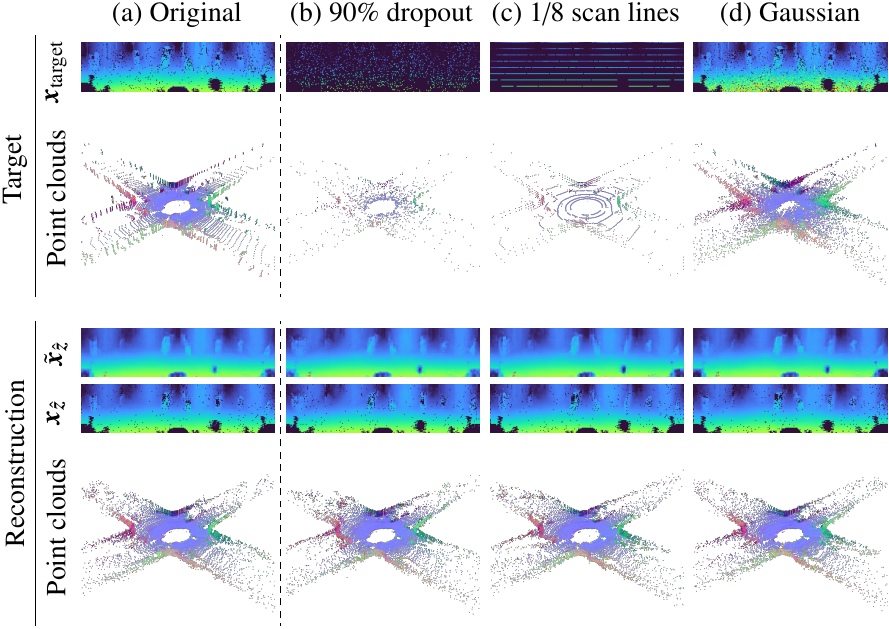}
	\caption{Reconstruction examples for the target LiDAR scans $\bm{x}_{\mathrm{target}}$ with three types of corruption. All results were generated by our DUSty-I trained on KITTI. (a) Sample without any corruption from the KITTI test set. (b) $90\%$ points are randomly dropped. (c) Only 8 out of 64 horizontal lines are observable. (d) An additive Gaussian noise from $\mathcal{N}(0, 0.01)$ is applied to a depth value.}
	\label{fig:examples:corruption}
\end{figure*}

In Fig.~\ref{fig:examples:corruption}, we show some examples from DUSty-I, where the targets are with three types of strong corruptions: 90\% of points are randomly dropped (Fig.~\ref{fig:examples:corruption}(b)), 8 out of 64 horizontal lines are observable (Fig.~\ref{fig:examples:corruption}(c)), and pixel-wise noise from $\mathcal{N}(0,0.01)$ are added to the depth values (Fig.~\ref{fig:examples:corruption}(d)).
Despite very sparse or noisy targets $\bm{x}_{\mathrm{target}}$, our method succeeded in reconstructing inverse depth maps with underlying complete surfaces $\tilde{\bm{x}}_{\hat{\bm{z}}}$ and rendering realistic point-drops.
On the other hand, we can also observe that the far objects and thin structures are still difficult to reconstruct in some cases.

\section{Conclusion}

In this paper, we proposed DUSty, a GAN framework for LiDAR scan synthesis based on the decomposed image representation of inverse depth and point-drop.
Our method showed effectiveness on both synthesis and reconstruction tasks compared to the baseline that directly models the LiDAR scans.
The reconstruction experiments indicated potential applications of our method, such as restoration of corrupted data and upsampling of sparse scans from low-cost LiDARs.
Our method would also be applicable to rendering realistic point-drops for fully measurable simulation data~\cite{wu2018squeezeseg, manivasagam2020lidarsim} since our pixel-wise depth reconstruction can simultaneously produce a measurability map.
Although this study used standard GAN architectures, further investigation on the network design could improve the generation quality.
Future work includes introducing our point-drop learning into the other large networks~\cite{karras2018progressive, karras2020analyzing} and applying our synthetic data to perception tasks.


\section*{Acknowledgment}

This work was supported by a Grant-in-Aid for JSPS Fellows Grant Number JP19J12159 and JSPS KAKENHI Grant Number JP20H00230.

%%%%%%%%%%%%%%%%%%%%%%%%%%%%%%%%%%%%%%%%%%%%%%%%%%%%%%%%%%%%%%%%%%%%%%%%%%%%%%%%

\bibliographystyle{IEEEtran}
\bibliography{main}

\end{document}